\pdfoutput=1

\documentclass[11pt]{article}

\usepackage[final]{coling}

\usepackage{times}
\usepackage{latexsym}

\usepackage[T1]{fontenc}

\usepackage[utf8]{inputenc}

\usepackage{microtype}

\usepackage{inconsolata}

\usepackage{graphicx}
\usepackage{booktabs}
\usepackage{multirow}
\usepackage{amsmath}
\usepackage{xcolor}

\title{Montague semantics and modifier consistency measurement in neural language models}

\author{\href{https://orcid.org/0000-0002-8124-9224}{Danilo S. Carvalho$^{1}$\includegraphics[scale=0.08]{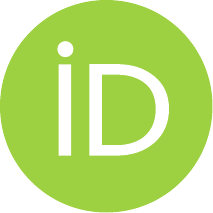}\hspace{1mm}},
~ \href{https://orcid.org/0000-0003-0028-5440}{Edoardo Manino$^{2}$\includegraphics[scale=0.08]{orcid.pdf}\hspace{1mm}},
~ \href{https://orcid.org/0000-0002-8668-3038}{Julia Rozanova$^{2}$\includegraphics[scale=0.08]{orcid.pdf}\hspace{1mm}},\\ 
\href{https://orcid.org/0000-0002-6235-4272}{\textbf{ Lucas Cordeiro$^{2}$}\includegraphics[scale=0.08]{orcid.pdf}\hspace{1mm}} \and 
\href{https://orcid.org/0000-0002-4430-4837}{\textbf{Andr\'{e} Freitas$^{1,2,3}$}\includegraphics[scale=0.08]{orcid.pdf}\hspace{1mm}} \\
National Biomarker Centre, CRUK-MI, Univ. of Manchester, United Kingdom$^{1}$\\
Department of Computer Science, University of Manchester, United Kingdom$^{2}$ \\
Idiap Research Institute, Switzerland$^{3}$ \\
\texttt{\{firstname.lastname\}@manchester.ac.uk}}

\begin{document}
\maketitle

\begin{abstract}
This work proposes a novel methodology for measuring compositional behavior in contemporary language embedding models. Specifically, we focus on adjectival modifier phenomena in adjective-noun phrases.
In recent years, distributional language representation models have demonstrated great practical success. At the same time, the need for interpretability has elicited questions on their intrinsic properties and capabilities. Crucially, distributional models are often inconsistent when dealing with compositional phenomena in natural language, which has significant implications for their \textit{safety} and \textit{fairness}. Despite this, most current research on compositionality is directed towards improving their performance on similarity tasks only. This work takes a different approach, introducing three novel tests of compositional behavior inspired by Montague semantics. Our experimental results indicate that current neural language models do not behave according to the expected linguistic theories. This indicates that current language models may lack the capability to capture the semantic properties we evaluated on limited context, or that linguistic theories from Montagovian tradition may not match the expected capabilities of distributional models.
\end{abstract}

\section{Introduction}

Distributional semantics and neural language models have been a dominant approach in language representation models for nearly a decade since the emergence of deep learning methods~\citep{lenci2022comparative}. This is due to the consistent achievements in terms of the state-of-the-art performance in various downstream NLP tasks and the progressive increase of their parameter size and complexity. Interest in the properties of these models and their relationships with semantic formalisms is older than their rise to mainstream use~\citep{baroni2010nouns}. However, the recent demand for models delivering safety guarantees and better inference control has highlighted its importance~\citep{floridi2020gpt}. This is of particular relevance to retrieval-augmented generation (RAG)~\cite{NEURIPS2020_6b493230}, as implicit compositional assumptions are a source of semantic gaps in natural language queries.

Indeed, understanding the intrinsic linguistic and semantic properties of distributional neural language models can provide important insight on their capabilities and limitations. 
From a purely distributional perspective, studies have been conducted on analysing the concept drift~\citep{sommerauer2019conceptual} and biases~\cite{bhardwaj2021investigating} of such models. On a linguistic front, attempts at mapping vector representations to dictionary senses and lexical features have yielded promising results~\citep{pilehvar2015senses, carvalho2017building}. Similarly, works that probed for the presence of linguistic features in sentence-level representations revealed a wide array of syntactic information captured~\citep{miaschi2020contextual, ferreira2021does}.

However, one issue that has been underexplored from a linguistic standpoint is compositionality and their associated set-theoretic (Montagovian) concepts, where efforts have been directed towards improving performance of the representations on similarity tasks (see Section~\ref{sec:related}), without attempting to relate the linguistic principles involved with compositional properties observed.

This work proposes to fill this research gap, electing the \textit{modifier phenomena}~\citep{dixon2004adjective, morzycki2016modification} as a starting point for the analysis of compositional properties in language models, and adopting text embeddings as proxies of concept denotations. In this way, we can test the manifestation of compositional properties in adjective phrase denotations, such as intersectivity, as a function of the consistency of geometric properties in the embedding space, in the form of \textit{metamorphic relations}~\citep{Chen2018}. The hypothesis of proxying denotations through embeddings has been implicitly used for the ``vector analogy'' tasks (e.g., $king - man + woman = queen$)~\citep{mikolov2013linguistic}, but is used here explicitly to test denotation properties in the embedding space.


Similarly, the concept of metamorphic relation has recently waded its way from the field of software engineering \citep{Chen2018} to machine learning and natural language processing \citep{Belinkov2018, manino2022}. There, it brings the promise of formally defining the expected behavior of a learning-based model and rigorously testing whether it holds in practice without the need for ground-truth labels. Popular applications of behavioral testing usually focus on plain substitutions of similar words (e.g., robustness to synonym replacement) \citep{Jia2019}, or semantic opposition (e.g., changing the gender of nouns) \citep{Ma2020}. However, efforts have been made to extend this framework to higher-level linguistic properties such as systematicity and transitivity \citep{manino2022}. The present work continues this line of research by grounding the concept of metamorphic relation onto the linguistic tradition of formal semantics.

\noindent
\textbf{Hypothesis [embedding-denotation analogy]:} Assume that the modifier phenomena is described by a Set representation/Montague semantics compositional model. We expect a large language model, which at the limit captures the distributional properties of an infinite corpus of utterances, to show empirical evidence of the formal properties of the modifier phenomena.

\noindent
\textbf{Research Questions:} In this paper, we restrict our inquiry to adjective modifiers and contemporary neural language models. In this setting, we can pose the following research questions:

\textbf{RQ1.} Adjective-noun composition is described in Montague semantics as a function mapping elements between two sets A $\rightarrow$ P corresponding to the properties satisfied by the individuals referred by each set (denotation). Can we expect to observe a correspondence of these theoretical linguistic properties in neural language models that operate on dense vector spaces?

\textbf{RQ2.} Existing neural language models are limited by their choices of the learning process (objective functions) and the language data available for model training. To what degree can we observe evidence of the compositional effect of adjective modifiers? Do contextual models differ from non-contextual ones in this regard?

\noindent
\textbf{Contributions:} We propose a methodology for measuring the presence of compositional behaviour in contemporary neural language models related to adjectival modifier phenomena in adjective-noun phrases, from a Montagovian formalism perspective. Our methodology \textit{translates a set-based formal semantic theory into metamorphic relations in embedding spaces} based on the cosine distance between embeddings (RQ1). Our results show that current neural language models do not behave consistently according to the linguistic theories with regard to the evaluated intersective property. In fact, there is no statistically significant difference between different adjective categories: the empirical behaviour we observe tends to be \textit{intersective} across all inputs and language models (RQ2). Additionally, we found that while large SOTA transformer models behave similarly to non-contextual models regarding intersectivity, when accounting for mean-pooling bias, they largely differ in terms of subsectivity, placing heavy emphasis on adjectives instead of nouns (RQ2). The results indicate that current language models may lack the capability to capture the semantic properties we evaluated on limited context, or that linguistic theories from Montagovian tradition may not match expected capabilities of distributional models. Finally, we make publicly available the developed experimental pipeline and dataset (44652 adjective-noun phrases) for reproducibility purposes\footnote{Experimental code and the full dataset are available at: \url{https://github.com/dscarvalho/modifiers_consistency}\label{fn:dataset}}.

\noindent
\textbf{Scope of the study:} The formal linguistic properties evaluated in this study are not sufficient nor intended to evaluate general compositionality, but are a fundamental part of a larger set of compositional phenomena, which includes verbal, nominal and even non-linguistic (e.g., arithmetic) composition. Additionally, the proposed methodology focuses on the model's distributional embedding spaces, rather than specific downstream tasks. This allows it to be applied for general assessment of current and future models' compositional behaviour, irrespective of task capability or specialisation.

The remainder of this paper is organized as follows: Section~\ref{sec:mod_phenomena} explains the linguistic grounding of this work in more detail, Section~\ref{sec:methodology} discusses our methodology, Section~\ref{sec:exp} reports our experimental setup and discusses its findings, Section~\ref{sec:related} presents the broader landscape of related works, and finally Section~\ref{sec:conclusion} summarizes our contribution and concludes with some final remarks.

\begin{figure*}[ht!]
\begin{center}
    \includegraphics[width=0.8\linewidth]{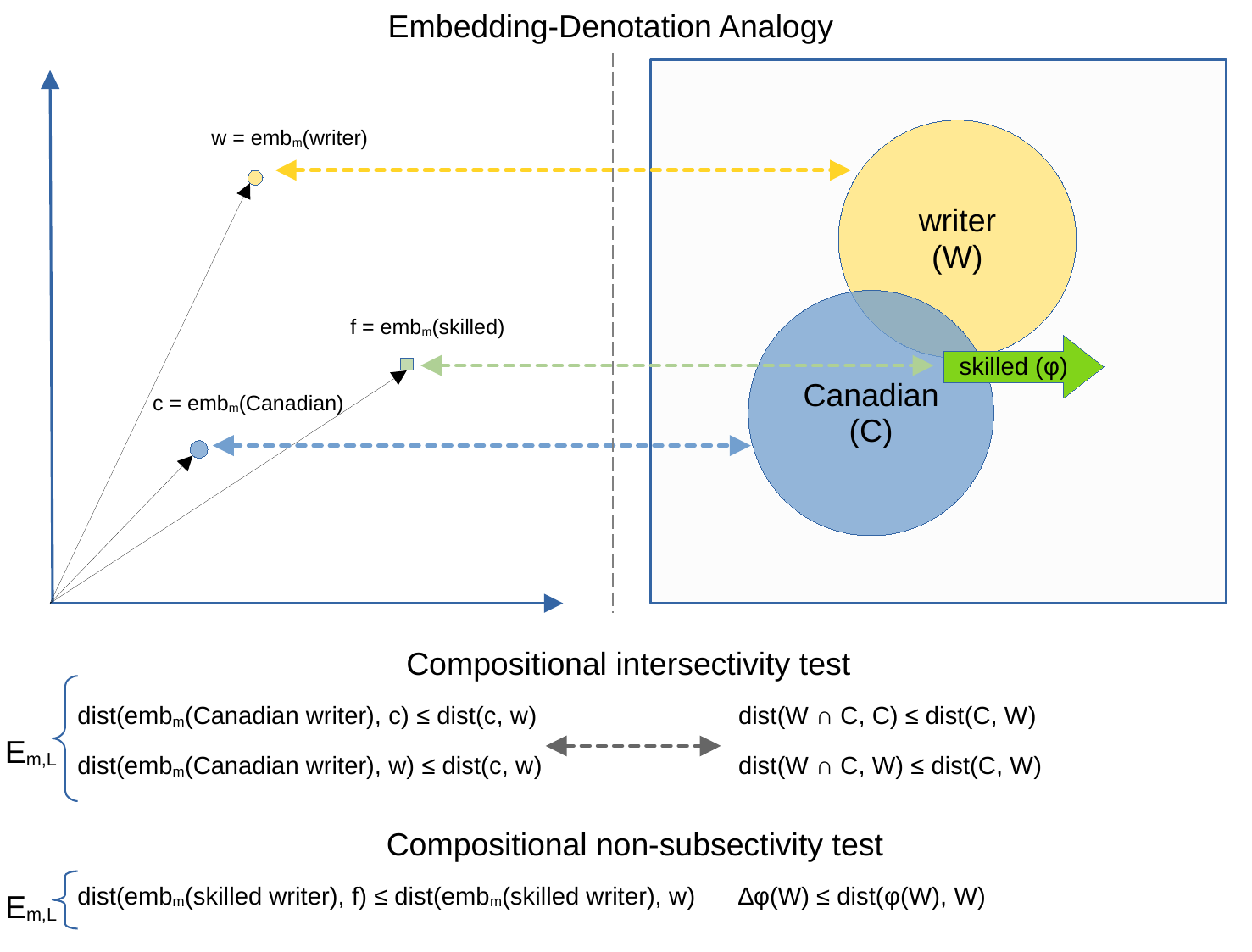}
    \caption{Methodology for testing model consistency regarding the modifier phenomena in adjective-noun phrases (ANs). $m$ represents a language model and $L$ the regular language $(adj~~)+noun$. $E_{m,L}$ is calculated by averaging the no. of combinations $(a, n, \phi, p = an)$ where the inequalities hold over the vocabulary size.}
    \label{fig:methodology}
\end{center}
\end{figure*}

\section{The modifier phenomena}
\label{sec:mod_phenomena}

Of all linguistic phenomena arising from the composition of meaning of two or more words, \textit{modification}, and in particular the application of adjectives, has been the subject of extensive study~\citep{dixon2004adjective, morzycki2016modification}.

\subsection{Modification semantics}

From a linguistic standpoint, \textit{modification} does not constitute a single grammatical phenomenon, being a term for expressions that do not fit into either the predicate or argument categories. In fact, \textit{modification} characterizes both a family of (internal) lexical semantic characteristics and of (external) distributional ones~\cite{morzycki2016modification}. For the purpose of our study, we narrow down the definition of modifiers to a set of compositional principles regarding intensional interpretations from a Montagovian formalism, with adjective phrases being the object of analysis \citep{boleda2013intensionality,paperno2016whole}. On the one hand, this choice allows us to interpret nouns as their denotations in set theoretical form. For example, we assume that the noun ``dog'' represents the set of properties that hold for any individual to which the concept of dog applies. On the other hand, this choice allows us to interpret adjectives as functions of set-based denotations. The latter is discussed below.

\subsection{Adjective types and interpretations}

Adjectives can be classified according to their effect on the denotations they modify \citep{morzycki2016modification,pavlick-callison-burch-2016-called}:

\begin{itemize}
    \item \textbf{Intersective} (or \textit{extensional}): describes the intersection of the noun denotation with the denotation of the adjective itself. Thus, the adjective can also be interpreted as a set. E.g. ``red car'' denotes the set of things that are both ``car'' and ``red''.
    
    \item \textbf{Subsective (non-intersective)}: describes a strict subset of the noun denotation it modifies. E.g. ``skillful teacher'' denotes a subset of teachers, but there is no general denotation for ``skillful''.
    
    \item \textbf{Privative non-subsective}: describes a set that is completely disjoint from the denotation of the noun it modifies. E.g. ``fake wall'' denotes a set of things that are definitely \textit{not} walls.
    
    \item \textbf{Plain non-subsective}: describes a set that may or may not be a subset of the noun denotation it modifies, depending on the context or the adjective itself. E.g. ``alleged criminal'' denotes a set of individuals whose inclusion in the set of criminals is dubious or undefined, while ``former president'' denotes a set of individuals that are not presidents anymore.
    
    \item \textbf{Ambiguous}: can be applied to any of the previous categories, depending on the context and the modified noun. E.g. ``big'' is intersective in the phrase ``big truck'' and subsective non-intersective in the phrase ``big fool''.
\end{itemize}

Section \ref{sec:set_based} contains further formalization of these adjective types and their related properties.

\subsection{Distributional questions}

Hanging fundamentally on the \textit{distributional hypothesis}, distributional models are primarily optimised for capturing statistical co-occurrence relations (syntagmatic and paradigmatic relations) at scale. As a result, distributional models naturally excel at computing measures of semantic \textit{relatedness} and semantic \textit{similarity} between any given pair of terms in a corpus. However, their ability at capturing more structured compositional behaviour is unclear.


Efforts at building distributional models that exhibit compositional behaviour \textit{by construction} has been made in the past \citep{clark2007combining,mitchell2008vector,guevara2010regression}. Unfortunately, these efforts predate the advent of state-of-the-art self-supervised language models, and cannot compete with their performance. In fact, recent language models have tackled composition in more implicit ways, with state-of-the-art approaches being trained on multiple objectives such as masked word prediction, sentence-level similarity and entailment functions \citep{reimers2019sentence,Sanh2019DistilBERTAD,ni2022sentence}.


This raises the question of whether the representations obtained in this way could be employed as \textit{proxies} for word and phrase denotations. If this is the case, then any term comparisons made in the embedding spaces would represent an equivalent operation between denotations (e.g., subset inclusion). Conversely, set theoretical properties on denotations could be interpreted as geometrical properties of the embedding space (e.g., vector distance constraints). This understanding lies at the foundation of the methodology presented hereon.


\section{Methodology}
\label{sec:methodology}

Our methodology is centred around the hypothesis that neural embeddings should correctly approximate the linguistic denotation of the input phrases. In this light, we propose three different metamorphic tests to check whether neural models satisfy such hypothesis.

\subsection{Set-based phrase denotations}
\label{sec:set_based}

In general, we say that a noun $n$ can be modified by an adjective $a$ to form an adjective-noun phrase $p=an$. The denotation of $p$ can be represented as a set, and depends on the type of the adjective $a$ (see Section~\ref{sec:mod_phenomena}). More specifically, we divide the adjectives into two main categories: \textit{intersective} and \textit{non-intersective}. Here, the \textit{non-intersective} category includes both subsective and non-subsective adjectives.

On the one hand, if $a$ is an intersective adjective, then the denotation of $p$ is simply the intersection of the denotations of $a$ and $n$. For example, the intersective phrase $p=Canadian\,writer$ is associated with the following Montague denotations (intensions):
\begin{equation*}
    \begin{aligned}
        n(x) &= \lambda x.[writer(x)] \\
        a(x) &= \lambda x.[Canadian(x)] \\
        p(x) &= \lambda x.[a(x) \land n(x)]
    \end{aligned}
\end{equation*}
and corresponding sets (extensions):
\begin{equation}
    \begin{aligned}
        \label{eq:set_intersective}
        N &\equiv \{x~|~n(x)=\top\} \\
        A &\equiv \{x~|~a(x)=\top\} \\
        P &\equiv A \cap N
    \end{aligned}
\end{equation}
where $P \subseteq N$ and $P \subseteq A$.

On the other hand, if $a$ is a non-intersective adjective, then the denotation of $p$ involves functions over sets. For example, the phrase $p=skilled\,writer$ requires the following Montague denotations:
\begin{equation*}
    \begin{aligned}
        a(n,x) &= \lambda n.\lambda x[skilled(n(x),x)] \\
        p(x) &= \lambda x.[a(W,x)]
    \end{aligned}
\end{equation*}
where function $a$ can discriminate whether $x$ is a skilled writer, but has no concept of ``skilfulness'' in general. Accordingly, the corresponding sets (extensions) are:
\begin{equation}
    \label{eq:set_non_intersective}
    P \equiv A \equiv \{x~|~p(x)=\top\} \subseteq N
\end{equation}

\begin{figure*}[ht!]
\begin{center}
    \includegraphics[width=0.92\linewidth]{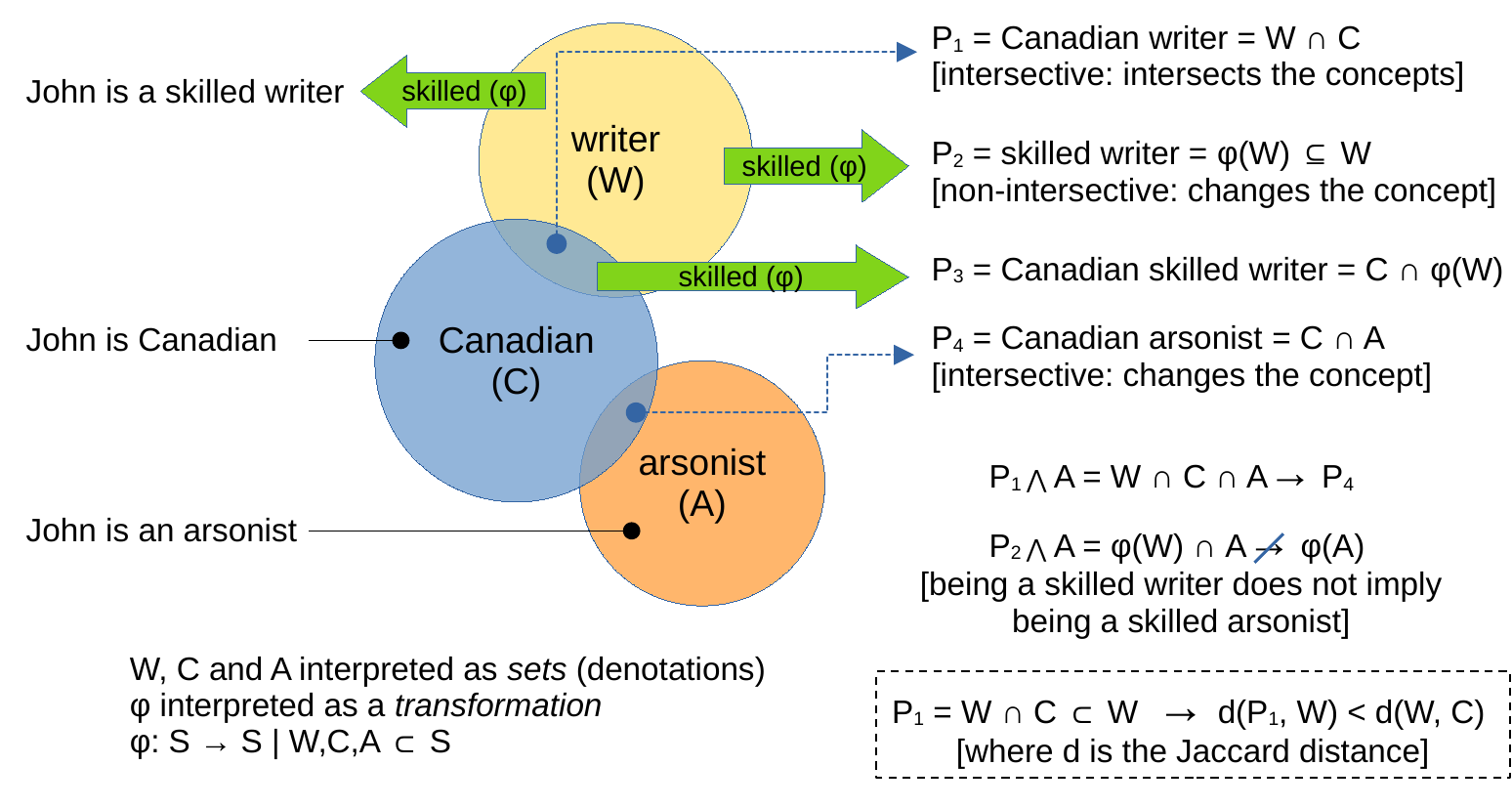}
    \caption{Intersective and non-intersective set relations in adjective-noun phrase denotations.}
    \label{fig:set_rel}
\end{center}
\end{figure*}

Note that in the intersective case (see Equation \ref{eq:set_intersective}) the set $P$ is included in both $A$ and $N$, whereas in the non-intersective case (see Equation \ref{eq:set_non_intersective}) this is not the case. As a result, if we could measure the distance between these three sets for a generic adjective-noun phrase $p=an$, then we should be able to identify the type of the adjective $a$. Figure~\ref{fig:set_rel} illustrates this concept of relations between sets.

\subsection{Embedding-denotation analogy}

Thus, our core hypothesis is the following. If the phrase embedding correctly represents its denotation, we should observe some analogous inclusion relations between them. Since embeddings are defined in vector space, the inclusion relations must be replaced with another appropriate measure (e.g., cosine, Euclidean). This hypothesis motivates the following tests.

\subsection{Testing intersectivity (single phrase)} \label{sec:intersect_test}

Assume $p=t_1t_2\dots t_h$ is an adjective-noun phrase containing one or more adjectives. If all adjectives were intersective, the corresponding set relations $P\subseteq T_i$ would be satisfied (see Section \ref{sec:set_based}). In contrast, any two individual terms $t_j,t_k$ are generally unrelated, yielding $T_j\not\subseteq T_k$. We hypothesise that set inclusion translates into shorter distances between embedding, which leads us to the following test of intersectivity:
\begin{equation}
    \begin{aligned}
        I_{m,p} &\equiv d(emb_m(p), emb_m(t_i)) \\
        &\leq d(emb_m(t_j), emb_m(t_k)) \\
        &\qquad\forall{i, j, k};~~ j < k
    \end{aligned}
    \label{eq:emb_rel}
\end{equation}
\noindent where $t_{1..h}$ is a term of the phrase $p$ and $emb_m$ is the embedding function for model $m$. We define the consistency of a model $m$ concerning Equation \ref{eq:emb_rel} by taking the expectation of its truth value:
\begin{equation}
\label{eq:expect_intersect}
    E_{m, L}\big\{I_{m,p} = \top\big\},
    \quad p \sim L
\end{equation}
where $L$ is the regular language ``$(adj~~)+noun$'' with alphabet $\Sigma$, and $p$ is extracted from $L$ according to a probability distribution.

\subsection{Testing intersectivity (phrase pairs)} \label{sec:intersect_test_pair}

Here, we give a more complex distance relation between pairs of adjective-noun phrases, which allows us to test for behaviour that result strictly from intersective effects, by controlling for the induced intersectivity from vector pooling while accounting for synonymy, if present. Define $p_{a_1n_1}$, $p_{a_1n_2}$, $p_{a_2n_1}$ and $p_{a_2n_2}$ as all possible concatenations of two adjectives $a_1 \neq a_2$ and two nouns $n_1 \neq n_2$. We expect the following metamorphic relation to holding:
\begin{equation}
\label{eq:pair_set_dist}
    \begin{aligned}
        II_{m,\{p\}} &= d(emb_m(p_{a_1n_1}), emb_m(p_{a_1n_2})) \\ 
        &\leq d(emb_m(p_{a_2n_1}), emb_m(p_{a_2n_2}))
    \end{aligned}
\end{equation}
when $a_1$ is intersective and $a_2$ is not. For example:
\begin{equation*}
    \begin{aligned}
        d(Canadian\,writer&,Canadian\,surgeon)\\
        \leq d(skillful\,writer&,skillful\,surgeon)
    \end{aligned}
\end{equation*}
\noindent where $Canadian$ is an intersective adjective and $skilful$ is not. This is because we expect a $Canadian\,writer$ to have something in common with a $Canadian\,surgeon$, i.e., the fact that they are both Canadian. In contrast, a $skilful\,writer$ and a $skilful\,surgeon$ are not similar as there is minimal overlap between their skills.

As for Equation \ref{eq:expect_intersect}, we call the consistency of $m$:
\begin{equation}
\label{eq:pair_consistency}
    E_{m, L^2}\big\{II_{m,\{p\}} = \top\big\},
    \quad \{p\} \sim L^2
\end{equation}
where $\{p\}\equiv\{a_1n_1,a_1n_2,a_2n_1,a_2n_2\}$. The value of Equation \ref{eq:pair_consistency} should approach 1.0 when all $a_1$ in $L^2$ are intersective and all $a_2$ are not.

\subsection{Testing non-subsectivity} \label{sec:nonsubsect_test}

We propose to test for non-subsectivity by looking at the relative change caused by an adjective to a noun when combined in a phrase. Let $p=an$ be an adjective-noun phrase with associated set $P$ and noun set $N$. Subsective composition guarantees $P \subseteq N$, whereas non-subsective composition does not. Consequently, we hypothesise that the embedding of $p$ is closer to $n$ when $a$ is subsective. Accordingly, we can test for non-subsectivity with the following metamorphic relation:
\begin{equation}
    \begin{aligned}
        NI_{m,p} &= d(emb_m(p), emb_m(a)) \\
         &\leq d(emb_m(p), emb_m(n))
    \end{aligned}
    \label{eq:emb_rel_ni}
\end{equation}
and the corresponding consistency metric:
\begin{equation}
\label{eq:expect_nonsubsect}
    E_{m, L}\big\{NI_{m,p} = \top\big\},
    \quad p \sim L
\end{equation}
where $L$ is the same language as in Equation \ref{eq:expect_intersect}.

\section{Experimentation and discussion}
\label{sec:exp}

\subsection{Experimental setup}

To perform the tests introduced in Section \ref{sec:methodology}, we need the following components: a measure of distance $d$ in embedding space, a set of adjective-noun phrases covering all adjective types (the input data), and the language models to be tested. In all our experiments we use the cosine distance, the input phrases and the language models described below.

\subsubsection{Data collection}

We consider adjective categories based on \citet{morzycki2016modification} and \citet{pavlick-callison-burch-2016-called}, where the latter provides a further subdivision of non-subsective adjectives. We select the examples in \cite{morzycki2016modification} for the sets of subsective adjectives, and we use the dataset in \cite{pavlick-callison-burch-2016-called} for the collection of non-subsective adjectives, both summing 61 adjectives. For each adjective in this initial list, a closest synonym was chosen for the phrase pair intersectivity test, totaling 122 adjectives. At the same time, we choose a set of 12 nouns covering both concrete and abstract concepts to form adjective/noun phrases.

\begin{table*}[ht!]
    \centering
    \begin{tabular}{@{}lllcl@{}}
    \toprule
    Adjective Type & Set-Theoretic Definition                                & Examples          & \# of Adjectives &  \\ \midrule
    Subsective (Intersective)     & $AN \subseteq N$ and $AN \subseteq A$                   & Red, Wild         & 22 &  \\
    Subsective (Non-Intersective) & $AN \subseteq N$ and $AN \not \subseteq A$              & Skilful, Rare     & 12 &  \\
    Non-Subsective (Plain)        & $AN \not \subseteq N$ and $AN \cap N \not = \emptyset $ & Alleged, Disputed & 54 &  \\
    Non-Subsective (Privative)    & $AN \cap N = \emptyset$                                 & Fake, Imaginary   & 28 &  \\
    Ambiguous                     & Contextually, one of the above                          & Old, Big          & 6 &  \\ \bottomrule
    \end{tabular}
    \caption{Adjective type composition for the vocabulary.}
    \label{tab:vocab_adj}
\end{table*}

The adjective and nouns lists were reviewed by each of the authors. While most adjective categorisations were left unchanged, we included an ``ambiguous'' category to house those adjectives that had ambiguous meaning within our phrase set. The categories and their definitions are presented in Table~\ref{tab:vocab_adj}. The complete list of categorised adjectives is included as supplementary material (Appendix~\ref{sec:apd_adj_list}).

\subsubsection{Phrase Generation}

The phrases were generated by using a regular language defined by the expression $(adj~)+noun$, where $adj$ and $noun$ are taken from the lists of adjectives and nouns respectively. More formally, $adj = (wild|red|...)$ and $noun = (student|dog|...)$. All the phrases up to 3 words were generated: e.g. ``wild dog'' and ``square assumed law''. The final dataset contains \textit{44652 phrases}\footref{fn:dataset}.

For reasons of space, we introduce a shorthand notation for the two types of phrases we generate: we write AN (respectively, AAN) to denote a phrase composed of a single adjective followed by a noun (respectively, two adjectives followed by a noun). With slight abuse of notation, we also use AN and AAN to refer to the set interpretation (denotation) of a phrase rather than the phrase itself.

\begin{table}[t!]
    \centering
    \resizebox{\columnwidth}{!}{
    \begin{tabular}{@{}llllll@{}}
    \toprule
    \multicolumn{1}{c}{\multirow{2}{*}{\textbf{Models}}} & \multicolumn{5}{c}{\textbf{Adjective Type}}                                                                                                                                                                                                                                                             \\
    \multicolumn{1}{c}{}                                 & \begin{tabular}[c]{@{}l@{}}S-I\end{tabular} & \begin{tabular}[c]{@{}l@{}}S-NI\end{tabular} & \begin{tabular}[c]{@{}l@{}}NS-Pl\end{tabular} & \begin{tabular}[c]{@{}l@{}}NS-Pr\end{tabular} & A \\ \midrule
    DPR                                                  & 0.86                                                                 & 0.90                                                                     & 0.85                                                              & 0.89                                                                  & 0.97       \\
    LaBSE                                                & 1.0                                                                  & 1.0                                                                      & 1.0                                                               & 1.0                                                                   & 1.0       \\
    Specter                                              & 0.93                                                                 & 0.99                                                                     & 0.97                                                              & 0.93                                                                  & 0.97       \\
    TE3-small                                            & 1.0                                                                  & 1.0                                                                      & 1.0                                                               & 1.0                                                                   & 1.0       \\ 
    NV-Embed-v2                                          & 0.73                                                                 & 0.67                                                                     & 0.8                                                               & 0.85                                                                  & 0.75       \\
    stella\_en\_1.5B\_v5                                 & 1.0                                                                  & 1.0                                                                      & 1.0                                                               & 1.0                                                                   & 1.0       \\ \midrule
    Glove                                                & 1.0                                                                  & 1.0                                                                      & 1.0                                                               & 1.0                                                                   & 1.0       \\
    Word2Vec                                             & 1.0                                                                  & 1.0                                                                      & 1.0                                                               & 1.0                                                                   & 1.0       \\ \bottomrule \\
    \end{tabular}%
    }
    \caption{Consistency scores of the intersective property in Equation~\ref{eq:expect_intersect}, for single adjective-noun phrases (AN format). We use the following shorthand notation in the columns:
    Ambiguous (A), Subsective-Intersective (S-I), Subsective Non-Intersective (S-NI), Plain Non-Subsective (NS-Pl), Privative Non-Subsective (NS-Pr).}
    \label{tab:set_dist_an}
    \end{table}

\subsubsection{Encoding Strategy \& Language Models}

As we investigate emergent compositional behaviour, we selected models that provide a single sentence representation rather than a sequence of token representations. Most often than not, these are variants of state-of-the-art transformer-based model. However, they are further trained to generate composed vector representations which are more informative than, for example, mean pooling of token representations. More specifically, we consider DPR~\cite{karpukhin2020dense}, LaBSE~\citep{feng2022language}, Specter~\cite{specter2020cohan}, OpenAI's text-embeddings-3-small [TE3-small]~\cite{openai_embeddings}, NV-Embed-v2~\cite{lee2024nv} and Stella[en\_1.5B\_v5]~\cite{stella_hf}. The last two being respectively the current first and third ranked at the MTEB benchmark~\cite{muennighoff2023mteb}. With the exception of TE3-small, which is closed-source, all selected transformer models compose token representations either by CLS hidden state pooling (DPR, LaBSE, Specter) or by a specialised attention model (NV-Embed-v2, Stella). This avoids the inherent intersective bias from mean pooling.

For comparison, we also run the experiments on non-contextual language models trained on a purely distributional objective. In particular, we use mean-pooled representations of the Word2Vec \citep{mikolov2013efficient} and Glove \citep{pennington2014glove} models. These models provide a useful baseline to compare the aforementioned contextual models against.

\subsection{Results and discussion}

\subsubsection{Intersectivity experiment (single phrase)}
\label{sec:exp_intersect}


\begin{table}[t]
\centering
    \resizebox{\columnwidth}{!}{
    \begin{tabular}{@{}llllllllll@{}}
    \toprule
    \multicolumn{1}{c}{\textbf{Models}} & \multicolumn{5}{c}{\textbf{Adjective Type Pair}}                                                                                                                                                                                                                                                                                                                                                                                                                                                                                               \\ \midrule
    \multicolumn{1}{c}{\textbf{}}       & \begin{tabular}[c]{@{}l@{}}(S-I,\\ S-I)\end{tabular}      & \begin{tabular}[c]{@{}l@{}}(S-NI,\\ S-I)\end{tabular}   & \begin{tabular}[c]{@{}l@{}}(NS-Pl, \\ S-I)\end{tabular}  & \begin{tabular}[c]{@{}l@{}}(NS-Pr, \\ S-I)\end{tabular}  & \begin{tabular}[c]{@{}l@{}}(A,\\ S-I)\end{tabular} \\ \midrule
    DPR                                 & 0.52                                                      & 0.43                                                     & 0.53                                                    & 0.52                                                    & 0.62                                             \\
    LaBSE                               & 0.92                                                      & 0.93                                                     & 0.95                                                    & 0.91                                                    & 0.97                                             \\
    Specter                             & 0.67                                                      & 0.73                                                     & 0.72                                                    & 0.67                                                    & 0.73                                             \\
    TE3-small                           & 1.0                                                       & 1.0                                                      & 1.0                                                     & 1.0                                                     & 1.0                                                \\ 
    NV-Embed-v2                         & 0.78                                                      & 0.71                                                     & 0.68                                                    & 0.81                                                    & 0.75                                               \\
    stella\_en\_1.5B\_v5                & 1.0                                                       & 1.0                                                      & 1.0                                                     & 1.0                                                     & 1.0                                                \\ \midrule
    Glove                               & 1.0                                                       & 1.0                                                      & 1.0                                                     & 0.94                                                    & 1.0                                                \\
    Word2Vec                            & 1.0                                                       & 1.0                                                      & 0.97                                                    & 0.94                                                    & 1.0                                                \\ \bottomrule \\
    \end{tabular}
    }
    \caption{Consistency scores of the intersective property in Equation~\ref{eq:expect_intersect}, for adjective-noun phrases with two adjectives (AAN format). Same notation as Table~\ref{tab:set_dist_an}. Results for all type combinations are included as supplementary material (Appendix~\ref{sec:apd_exp_results}).}
    \label{tab:set_dist_aan}
\end{table}

Our first metamorphic property from Section \ref{sec:intersect_test} requires that the embedding of an adjective-noun phrase lies closer to each term than the distance between any pair of terms. The results in Table~\ref{tab:set_dist_an} indicate that except for DPR, Specter and NV-Embed-v2, the models' pooling operations are equivalent to mean pooling, making them universally intersective. If we interpret the metric in Equation~\ref{eq:expect_intersect} as indicative of intersective behaviour, we would have to conclude that the remaining models do not behave according to the expected modifier phenomena formalisms, with higher consistency scores on non-intersective pairings. This conclusion is corroborated when we consider the results on phrases with two adjectives (AAN) in Table~\ref{tab:set_dist_aan}, which further highlights the differences between the models' compositional properties. Those can be summarised in the following findings:

\paragraph{Models with mean-pooling equivalent composition are universally intersective (vice-versa)}~\\
As averaging embeddings will always produce one that is the closest to both, satisfying Equation~\ref{eq:emb_rel}. Conversely, a universally intersective model is likely to have mean-pooling equivalent composition. This can be observed on LaBSE, TE3-small and Stella, which are not mean-pooling models.

\paragraph{Models without mean-pooling equivalent composition do not consistently capture adjective intersectivity}~\\
As observed on DPR, Specter and NV-Embed-v2, the embedding distance relations are dependent on attention parameters not corresponding with the adjective categorisation.

\noindent More details are available on Appendix~\ref{sec:apd_exp_results}.

\subsubsection{Intersectivity experiment (phrase pair)}
\label{sec:exp_intersect_2}

\begin{table}[t]
    \centering
    \resizebox{\columnwidth}{!}{
    \begin{tabular}{@{}llllllllll@{}}
    \toprule
    \multicolumn{1}{c}{\textbf{Models}} & \multicolumn{5}{c}{\textbf{Adjective Type Pair}}                                                                                                                                                                                                                                                                                                                                                                                                                                                                                               \\ \midrule
    \multicolumn{1}{c}{\textbf{}}       & \begin{tabular}[c]{@{}l@{}}(S-I,\\ S-I)\end{tabular}     & \begin{tabular}[c]{@{}l@{}}(S-I,\\ S-NI)\end{tabular}  & \begin{tabular}[c]{@{}l@{}}(S-I, \\ NS-Pl)\end{tabular} & \begin{tabular}[c]{@{}l@{}}(S-I, \\ NS-Pr)\end{tabular}  & \begin{tabular}[c]{@{}l@{}}(S-I,\\ A)\end{tabular}  \\ \midrule
    DPR                                 & 0.50                                                     & 0.32                                                   & 0.34                                                    & 0.50                                                    & 0.42                                                  \\
    LaBSE                               & 0.50                                                     & 0.42                                                   & 0.34                                                    & 0.53                                                    & 0.33                                                  \\
    Specter                             & 0.50                                                     & 0.65                                                   & 0.55                                                    & 0.50                                                    & 0.57                                                  \\
    TE3-small                           & 0.50                                                     & 0.51                                                   & 0.48                                                    & 0.48                                                    & 0.82                                                  \\
    NV-Embed-v2                         & 0.50                                                     & 0.54                                                   & 0.51                                                    & 0.51                                                    & 0.82                                                  \\
    stella\_en\_1.5B\_v5                & 0.50                                                     & 0.75                                                   & 0.64                                                    & 0.58                                                    & 0.91                                                  \\ \midrule
    Glove                               & 0.50                                                     & 0.66                                                   & 0.69                                                    & 0.70                                                    & 0.47                                                  \\
    Word2Vec                            & 0.50                                                     & 0.75                                                   & 0.65                                                    & 0.49                                                    & 1.0                                                   \\ \bottomrule \\
    \end{tabular}
    }
    \caption{Consistency score of the intersective property in Equation~\ref{eq:pair_consistency}, for pairs of adjective-noun phrases with a single adjective (AN format). Same notation as Table~\ref{tab:set_dist_an}.}
    \label{tab:setdist_pair_an}
\end{table}

Our second metamorphic property from Section \ref{sec:intersect_test_pair} completes the picture on intersectivity. The property requires adjective-noun phrases that share the same intersective adjective to be closer to each other than phrases with non-intersective ones. Table~\ref{tab:setdist_pair_an} reports the results of our experiments, which suggest that each model places intersective emphasis in a different category of adjectives, with Stella being the one that most closely approaches the linguistically expected behaviour, together with the non-contextual baselines. 

\subsubsection{Non-subsectivity experiment} \label{sec:exp_nonsubsect}

Our third metamorphic relation from Section \ref{sec:nonsubsect_test} requires the adjective to ``pull'' the embedding of the whole phrase closer to them than the associated noun. This is a reasonable requirement because non-subsective adjectives completely change the meaning of the noun, rather than just specializing it. Our final experiment allows us to test whether this is indeed the behaviour of contemporary language models.


The results are in in Table~\ref{tab:non-sub}. Here, there is a clear trend regarding the size of the models, with the larger ones (in order of size: Stella, NV-Embed-v2, TE3-small) having substantially higher consistency scores overall. This indicates that those models place a much larger weight on the adjectives than the nouns in the composition process. The key findings of this experiment can be summarised as follows:

\paragraph{None of the tested models behave according to the expectations given by the subsectivity formalism}~\\
The consistency scores show different patterns of subsectivity w.r.t. the categories for each model, but in none of them the highest score belongs to a `NS' category.

\paragraph{Larger models composition process largely emphasises adjectives instead of nouns}~\\
This effect is mostly independent of the adjective category, with the exception of ambiguous ones, which can be observed in TE3-small and Stella.\\

A case of particular interest is the ambiguously typed adjectives (dependent on the represented word sense): we see that the models do not always seem to agree on the chosen sense. The numerical behaviour hints at whether the model is more likely to choose intersective or non-intersective senses of adjectives such as "old".

Thus, adjective type differences display relatively low compositional effects on broad intersectivity and subsectivity in the evaluated models. This phenomenon indicates that while such differences may be encoded in individual word representations, specially on non-contextual models, they do not transfer generally in the expected way to the compositions. 

\begin{table}[t!]
\resizebox{\columnwidth}{!}{
\begin{tabular}{@{}llllllllll@{}}
\toprule
\multicolumn{1}{c}{\multirow{2}{*}{\textbf{Models}}} & \multicolumn{5}{c}{\textbf{Adjective Type}}                                                                                                                                                                                                                                                                                                                                                                 \\
\multicolumn{1}{c}{}                                 & \begin{tabular}[c]{@{}c@{}}S-I\end{tabular} & \begin{tabular}[c]{@{}c@{}}S-NI\end{tabular} & \begin{tabular}[c]{@{}c@{}}NS-Pl\end{tabular} & \begin{tabular}[c]{@{}c@{}}NS-Pr\end{tabular} & \begin{tabular}[c]{@{}c@{}}A\end{tabular} \\ \midrule
DPR                                                  & 0.46                                                            & 0.37                                                             & 0.48                                                             & 0.54                                                             & 0.39       \\
LaBSE                                                & 0.36                                                            & 0.31                                                             & 0.51                                                             & 0.33                                                             & 0.19       \\
Specter                                              & 0.48                                                            & 0.31                                                             & 0.49                                                             & 0.57                                                             & 0.33   \\
TE3-small                                            & 0.81                                                            & 0.75                                                             & 0.74                                                             & 0.77                                                             & 0.39  \\
NV-Embed-v2                                          & 0.84                                                            & 0.79                                                             & 0.79                                                             & 0.83                                                             & 0.81   \\ 
stella\_en\_1.5B\_v5                                 & 0.81                                                            & 0.56                                                             & 0.58                                                             & 0.64                                                             & 0.33       \\ \midrule
Glove                                                & 0.61                                                            & 0.22                                                             & 0.22                                                             & 0.32                                                             & 0.28      \\
Word2Vec                                             & 0.55                                                            & 0.21                                                             & 0.34                                                             & 0.49                                                             & 0.0       \\ \bottomrule \\
\end{tabular}%
}
\caption{Consistency score (consistency) of the non-subsective property in Equation~\ref{eq:expect_nonsubsect}, for single adjective-noun phrases (AN format). Same notation as Table~\ref{tab:set_dist_an}.}
\label{tab:non-sub}
\end{table}

\section{Related work}
\label{sec:related}

Before the advent of self-supervised language models, much work has gone into constructing formally-motivated vector representations. To this end, \citet{clark2007combining} employs tensor product operations composition, while \citep{clark2008compositional} complements the previous approach with pregroup semantics. Similarly, \citet{mitchell2008vector} employs vector sums and products, whereas \citet{guevara2010regression}, \citet{guevara2011computing} and \citet{baroni2010nouns} model composition as a learnable function of two vectors. In the same vein, \citet{paperno2014practical} proposes a generalised representation of composition functions.


At the same time, existing studies cover a wide range of modifier phenomena: adjective-noun (AN) compositions \citep{boleda2013intensionality}, verb-argument composition \citep{lenci2011composing}, determiner-noun (DP) phrases \citep{bernardi2013relatedness}, recursive adjectival modifications \citep{vecchi2013studying}, reverse adjectival composition for phrase generation \citep{dinu2014make}, pointwise mutual information (PMI) analysis over AN compositions \citep{paperno2016whole}, morpheme representation \citep{marelli2015affixation} and metaphorical sense modeling \citep{lazaridou2013compositional,gutierrez2016literal}.

More recently, syntax-aware composition of dependency tree nodes is comprehensively addressed by \citet{weir2016aligning}, with empirical results tying previous approaches together. This work is complemented by \citet{gamallo2021compositional} using contextual representations from transformer models. Finally, \citet{purver2021incremental} proposes a dynamic syntax framework for unambiguous composition of sentences through incremental semantic parsing, which was evaluated with non-contextual representations.

After the advent of contextual transformer-based representation, the interest has shifted into \emph{testing} for specific compositional behaviours. 
The majority of existing works on \emph{metamorphic} testing of language models focus on checking simple behavioural rules at scale \citep{Belinkov2018}. This procedure is sometimes referred to as \emph{behavioural} testing, as in \citep{Ribeiro2020}.

For example, \citet{Ma2020} investigate fairness-related behaviours by measuring the model robustness to changes in the gender of nouns or addition of population-specific adjectives. Similarly, \citet{Sun2018} focus on multi-language machine translation and compare direct translations to multi-hop ones. Likewise, \citet{Tu2021} test the robustness of question-answer systems to changes in the given text.
Finally, \citet{manino2022} define higher-order metamorphic relations that simultaneously mutate multiple base inputs. Thanks to this, they can test the systematicity and transitivity of language models.


Our work centers on behavioural testing for an embedding-denotation analogy, attempting to address concerns (in the fresh context of contextual transformer-based representations) such as the limitation stated in \cite{kartsaklis2014compositional}: that compositions may describe spurious relations which result from expressiveness limitations, rather than modelling theoretical compositional behaviour.


\section{Conclusion} \label{sec:conclusion}
In this paper, we presented a methodology for measuring the presence and consistency of compositional behaviour in existing language models (LMs), comprising a set of tests for consistency of metamorphic relations associated to adjectival modifier phenomena in adjective-noun phrases, from a Montagovian formalism perspective. Our approach can provide important insight on LMs capabilities and limitations beyond semantic relatedness/similarity, helping to shape expectations on their use in applications with higher safety/criticality/fairness requirements. Although the tests are limited in scope, they can be applied to any language embedding.

Our empirical evaluation results indicate that current neural language models do not behave consistently according to expected behavior from the formalisms, with regard to the evaluated intersective and subsective properties. Such results imply that current language models, given limited context, may not be capable of capturing the evaluated semantic properties of language, or that linguistic theories from Montagovian tradition are not matching the expected capabilities of distributional models.

The proposed methodology is intended to be a stepping stone which can pave the way to a better understanding of LLMs latent spaces. Nevertheless, to improve our understanding of LLMs compositional capabilities, it is necessary to examine the relationship between the observed representation properties and specific NLP downstream task performance. Specifically, the alignment of compositional semantics between inputs and expected outputs, e.g., in RAG, summarization or Question Answering. Those should be explored in future work.

Future work also includes expanding the scope of the tests to other linguistic properties and an investigation on the effect of measured consistency on the relevant downstream tasks (e.g., NLI, RAG).

\section*{Limitations}

Having been designed as a set of measurements for quasi-symbolic analogy, the presented approach is not intended to demonstrate or prove the properties of the distributional models but rather to verify compliance to particular behaviours of interest.

The formal linguistic properties evaluated in this study are not sufficient nor intended to evaluate general compositionality, but are a fundamental part of a larger set of compositional phenomena, which includes verbal, nominal and even non-linguistic (e.g., arithmetic) composition.

Furthermore, while the Montagovian perspective of compositionality is highly relevant from the symbolic and verification standpoints, other theoretical frameworks can present different constraints regarding word and phrase interpretations and are worthy of exploration.

\section*{Acknowledgements}
This work was partially funded by the EPSRC grant EP/T026995/1 entitled “EnnCore: End-to-End Conceptual Guarding of Neural Architectures” under Security for all in an AI enabled society, by the Swiss National Science Foundation (SNSF) project NeuMath (\href{https://data.snf.ch/grants/grant/204617}{200021\_204617}), by the CRUK National Biomarker Centre, and supported by the Manchester Experimental Cancer Medicine Centre and the NIHR Manchester Biomedical Research Centre.

\bibliography{references}

\begin{thebibliography}{53}
\providecommand{\natexlab}[1]{#1}

\bibitem[{Baroni and Zamparelli(2010)}]{baroni2010nouns}
Marco Baroni and Roberto Zamparelli. 2010.
\newblock Nouns are vectors, adjectives are matrices: Representing adjective-noun constructions in semantic space.
\newblock In \emph{Proceedings of the 2010 conference on empirical methods in natural language processing}, pages 1183--1193.

\bibitem[{Belinkov and Bisk(2018)}]{Belinkov2018}
Yonatan Belinkov and Yonatan Bisk. 2018.
\newblock \href {https://openreview.net/forum?id=BJ8vJebC-} {Synthetic and natural noise both break neural machine translation}.
\newblock In \emph{International Conference on Learning Representations}.

\bibitem[{Bernardi et~al.(2013)Bernardi, Dinu, Marelli, and Baroni}]{bernardi2013relatedness}
Raffaella Bernardi, Georgiana Dinu, Marco Marelli, and Marco Baroni. 2013.
\newblock A relatedness benchmark to test the role of determiners in compositional distributional semantics.
\newblock In \emph{Proceedings of the 51st Annual Meeting of the Association for Computational Linguistics (Volume 2: Short Papers)}, pages 53--57.

\bibitem[{Bhardwaj et~al.(2021)Bhardwaj, Majumder, and Poria}]{bhardwaj2021investigating}
Rishabh Bhardwaj, Navonil Majumder, and Soujanya Poria. 2021.
\newblock Investigating gender bias in bert.
\newblock \emph{Cognitive Computation}, 13(4):1008--1018.

\bibitem[{Boleda et~al.(2013)Boleda, Baroni, McNally et~al.}]{boleda2013intensionality}
Gemma Boleda, Marco Baroni, Louise McNally, et~al. 2013.
\newblock Intensionality was only alleged: On adjective-noun composition in distributional semantics.
\newblock In \emph{Proceedings of the 10th International Conference on Computational Semantics (IWCS 2013): long papers; 2013 Mar 20-22; Postdam, Germany. Stroudsburg (USA): Association for Computational Linguistics (ACL); 2013. p. 35-46.} ACL (Association for Computational Linguistics).

\bibitem[{Carvalho and Nguyen(2017)}]{carvalho2017building}
Danilo~Silva Carvalho and Minh~Le Nguyen. 2017.
\newblock Building lexical vector representations from concept definitions.
\newblock In \emph{Proceedings of the 15th Conference of the European Chapter of the Association for Computational Linguistics: Volume 1, Long Papers}, pages 905--915.

\bibitem[{Chen et~al.(2018)Chen, Kuo, Liu, Poon, Towey, Tse, and Zhou}]{Chen2018}
Tsong~Yueh Chen, Fei-Ching Kuo, Huai Liu, Pak-Lok Poon, Dave Towey, T.~H. Tse, and Zhi~Quan Zhou. 2018.
\newblock \href {https://doi.org/10.1145/3143561} {Metamorphic testing: A review of challenges and opportunities}.
\newblock \emph{ACM Comput. Surv.}, 51(1).

\bibitem[{Clark et~al.(2008)Clark, Coecke, and Sadrzadeh}]{clark2008compositional}
Stephen Clark, Bob Coecke, and Mehrnoosh Sadrzadeh. 2008.
\newblock A compositional distributional model of meaning.
\newblock In \emph{Proceedings of the Second Quantum Interaction Symposium (QI-2008)}, pages 133--140. Oxford.

\bibitem[{Clark and Pulman(2007)}]{clark2007combining}
Stephen Clark and Stephen Pulman. 2007.
\newblock Combining symbolic and distributional models of meaning.
\newblock In \emph{Proceedings of AAAI Spring Symposium on Quantum Interaction}.

\bibitem[{Cohan et~al.(2020)Cohan, Feldman, Beltagy, Downey, and Weld}]{specter2020cohan}
Arman Cohan, Sergey Feldman, Iz~Beltagy, Doug Downey, and Daniel~S. Weld. 2020.
\newblock {SPECTER: Document-level Representation Learning using Citation-informed Transformers}.
\newblock In \emph{ACL}.

\bibitem[{Dinu and Baroni(2014)}]{dinu2014make}
Georgiana Dinu and Marco Baroni. 2014.
\newblock How to make words with vectors: Phrase generation in distributional semantics.
\newblock In \emph{Proceedings of the 52nd Annual Meeting of the Association for Computational Linguistics (Volume 1: Long Papers)}, pages 624--633.

\bibitem[{Dixon et~al.(2004)Dixon, Aikhenvald, A{\u\i}khenval'd, and Aikhenvald}]{dixon2004adjective}
R.M.W. Dixon, A.I. Aikhenvald, A.I.U. A{\u\i}khenval'd, and A.Y. Aikhenvald. 2004.
\newblock \href {https://books.google.co.uk/books?id=S64UDAAAQBAJ} {\emph{Adjective Classes: A Cross-Linguistic Typology}}.
\newblock Explorations in Language and S. OUP Oxford.

\bibitem[{Feng et~al.(2022)Feng, Yang, Cer, Arivazhagan, and Wang}]{feng2022language}
Fangxiaoyu Feng, Yinfei Yang, Daniel Cer, Naveen Arivazhagan, and Wei Wang. 2022.
\newblock Language-agnostic bert sentence embedding.
\newblock In \emph{Proceedings of the 60th Annual Meeting of the Association for Computational Linguistics (Volume 1: Long Papers)}, pages 878--891.

\bibitem[{Ferreira et~al.(2021)Ferreira, Rozanova, Thayaparan, Valentino, and Freitas}]{ferreira2021does}
Deborah Ferreira, Julia Rozanova, Mokanarangan Thayaparan, Marco Valentino, and Andr{\'e} Freitas. 2021.
\newblock Does my representation capture x? probe-ably.
\newblock In \emph{Proceedings of the 59th Annual Meeting of the Association for Computational Linguistics and the 11th International Joint Conference on Natural Language Processing: System Demonstrations}, pages 194--201.

\bibitem[{Floridi and Chiriatti(2020)}]{floridi2020gpt}
Luciano Floridi and Massimo Chiriatti. 2020.
\newblock Gpt-3: Its nature, scope, limits, and consequences.
\newblock \emph{Minds and Machines}, 30(4):681--694.

\bibitem[{Gamallo(2021)}]{gamallo2021compositional}
Pablo Gamallo. 2021.
\newblock Compositional distributional semantics with syntactic dependencies and selectional preferences.
\newblock \emph{Applied Sciences}, 11(12):5743.

\bibitem[{Guevara(2010)}]{guevara2010regression}
Emiliano~Raul Guevara. 2010.
\newblock A regression model of adjective-noun compositionality in distributional semantics.
\newblock In \emph{Proceedings of the 2010 workshop on geometrical models of natural language semantics}, pages 33--37.

\bibitem[{Guevara(2011)}]{guevara2011computing}
Emiliano~Raul Guevara. 2011.
\newblock Computing semantic compositionality in distributional semantics.
\newblock In \emph{Proceedings of the Ninth International Conference on Computational Semantics (IWCS 2011)}.

\bibitem[{Gutierrez et~al.(2016)Gutierrez, Shutova, Marghetis, and Bergen}]{gutierrez2016literal}
E~Dario Gutierrez, Ekaterina Shutova, Tyler Marghetis, and Benjamin Bergen. 2016.
\newblock Literal and metaphorical senses in compositional distributional semantic models.
\newblock In \emph{Proceedings of the 54th Annual Meeting of the Association for Computational Linguistics (Volume 1: Long Papers)}, pages 183--193.

\bibitem[{[@HuggingFace](2024)}]{stella_hf}
DunZhang [@HuggingFace]. 2024.
\newblock Stella-en-1.5b-v5.
\newblock \url{https://huggingface.co/dunzhang/stella_en_1.5B_v5}.

\bibitem[{Jia et~al.(2019)Jia, Raghunathan, G{\"o}ksel, and Liang}]{Jia2019}
Robin Jia, Aditi Raghunathan, Kerem G{\"o}ksel, and Percy Liang. 2019.
\newblock \href {https://doi.org/10.18653/v1/D19-1423} {Certified robustness to adversarial word substitutions}.
\newblock In \emph{Proceedings of the 2019 Conference on Empirical Methods in Natural Language Processing and the 9th International Joint Conference on Natural Language Processing (EMNLP-IJCNLP)}, pages 4129--4142, Hong Kong, China. Association for Computational Linguistics.

\bibitem[{Karpukhin et~al.(2020)Karpukhin, Oguz, Min, Lewis, Wu, Edunov, Chen, and Yih}]{karpukhin2020dense}
Vladimir Karpukhin, Barlas Oguz, Sewon Min, Patrick Lewis, Ledell Wu, Sergey Edunov, Danqi Chen, and Wen-tau Yih. 2020.
\newblock \href {https://doi.org/10.18653/v1/2020.emnlp-main.550} {Dense passage retrieval for open-domain question answering}.
\newblock In \emph{Proceedings of the 2020 Conference on Empirical Methods in Natural Language Processing (EMNLP)}, pages 6769--6781, Online. Association for Computational Linguistics.

\bibitem[{Kartsaklis(2014)}]{kartsaklis2014compositional}
Dimitri Kartsaklis. 2014.
\newblock Compositional operators in distributional semantics.
\newblock \emph{Springer Science Reviews}, 2(1):161--177.

\bibitem[{Lazaridou et~al.(2013)Lazaridou, Marelli, Zamparelli, and Baroni}]{lazaridou2013compositional}
Angeliki Lazaridou, Marco Marelli, Roberto Zamparelli, and Marco Baroni. 2013.
\newblock Compositional-ly derived representations of morphologically complex words in distributional semantics.
\newblock In \emph{Proceedings of the 51st Annual Meeting of the Association for Computational Linguistics (Volume 1: Long Papers)}, pages 1517--1526.

\bibitem[{Lee et~al.(2024)Lee, Roy, Xu, Raiman, Shoeybi, Catanzaro, and Ping}]{lee2024nv}
Chankyu Lee, Rajarshi Roy, Mengyao Xu, Jonathan Raiman, Mohammad Shoeybi, Bryan Catanzaro, and Wei Ping. 2024.
\newblock Nv-embed: Improved techniques for training llms as generalist embedding models.
\newblock \emph{arXiv preprint arXiv:2405.17428}.

\bibitem[{Lenci(2011)}]{lenci2011composing}
Alessandro Lenci. 2011.
\newblock Composing and updating verb argument expectations: A distributional semantic model.
\newblock In \emph{Proceedings of the 2nd workshop on cognitive modeling and computational linguistics}, pages 58--66.

\bibitem[{Lenci et~al.(2022)Lenci, Sahlgren, Jeuniaux, Cuba~Gyllensten, and Miliani}]{lenci2022comparative}
Alessandro Lenci, Magnus Sahlgren, Patrick Jeuniaux, Amaru Cuba~Gyllensten, and Martina Miliani. 2022.
\newblock A comparative evaluation and analysis of three generations of distributional semantic models.
\newblock \emph{Language Resources and Evaluation}, pages 1--45.

\bibitem[{Lewis et~al.(2020)Lewis, Perez, Piktus, Petroni, Karpukhin, Goyal, K\"{u}ttler, Lewis, Yih, Rockt\"{a}schel, Riedel, and Kiela}]{NEURIPS2020_6b493230}
Patrick Lewis, Ethan Perez, Aleksandra Piktus, Fabio Petroni, Vladimir Karpukhin, Naman Goyal, Heinrich K\"{u}ttler, Mike Lewis, Wen-tau Yih, Tim Rockt\"{a}schel, Sebastian Riedel, and Douwe Kiela. 2020.
\newblock \href {https://proceedings.neurips.cc/paper_files/paper/2020/file/6b493230205f780e1bc26945df7481e5-Paper.pdf} {Retrieval-augmented generation for knowledge-intensive nlp tasks}.
\newblock In \emph{Advances in Neural Information Processing Systems}, volume~33, pages 9459--9474. Curran Associates, Inc.

\bibitem[{Ma et~al.(2020)Ma, Wang, and Liu}]{Ma2020}
Pingchuan Ma, Shuai Wang, and Jin Liu. 2020.
\newblock \href {https://doi.org/10.24963/ijcai.2020/64} {Metamorphic testing and certified mitigation of fairness violations in {NLP} models}.
\newblock In \emph{Proceedings of the Twenty-Ninth International Joint Conference on Artificial Intelligence, {IJCAI} 2020}, pages 458--465. ijcai.org.

\bibitem[{Manino et~al.(2022)Manino, Rozanova, Carvalho, Freitas, and Cordeiro}]{manino2022}
Edoardo Manino, Julia Rozanova, Danilo Carvalho, Andre Freitas, and Lucas Cordeiro. 2022.
\newblock \href {https://doi.org/10.18653/v1/2022.findings-acl.185} {Systematicity, compositionality and transitivity of deep {NLP} models: a metamorphic testing perspective}.
\newblock In \emph{Findings of the Association for Computational Linguistics: ACL 2022}, pages 2355--2366, Dublin, Ireland. Association for Computational Linguistics.

\bibitem[{Marelli and Baroni(2015)}]{marelli2015affixation}
Marco Marelli and Marco Baroni. 2015.
\newblock Affixation in semantic space: Modeling morpheme meanings with compositional distributional semantics.
\newblock \emph{Psychological review}, 122(3):485.

\bibitem[{Miaschi and Dell’Orletta(2020)}]{miaschi2020contextual}
Alessio Miaschi and Felice Dell’Orletta. 2020.
\newblock Contextual and non-contextual word embeddings: an in-depth linguistic investigation.
\newblock In \emph{Proceedings of the 5th Workshop on Representation Learning for NLP}, pages 110--119.

\bibitem[{Mikolov et~al.(2013{\natexlab{a}})Mikolov, Chen, Corrado, and Dean}]{mikolov2013efficient}
Tomas Mikolov, Kai Chen, Greg Corrado, and Jeffrey Dean. 2013{\natexlab{a}}.
\newblock Efficient estimation of word representations in vector space.
\newblock \emph{arXiv preprint arXiv:1301.3781}.

\bibitem[{Mikolov et~al.(2013{\natexlab{b}})Mikolov, Yih, and Zweig}]{mikolov2013linguistic}
Tom{\'a}{\v{s}} Mikolov, Wen-tau Yih, and Geoffrey Zweig. 2013{\natexlab{b}}.
\newblock Linguistic regularities in continuous space word representations.
\newblock In \emph{Proceedings of the 2013 conference of the north american chapter of the association for computational linguistics: Human language technologies}, pages 746--751.

\bibitem[{Mitchell and Lapata(2008)}]{mitchell2008vector}
Jeff Mitchell and Mirella Lapata. 2008.
\newblock Vector-based models of semantic composition.
\newblock In \emph{proceedings of ACL-08: HLT}, pages 236--244.

\bibitem[{Morzycki(2016)}]{morzycki2016modification}
M.~Morzycki. 2016.
\newblock \href {https://books.google.co.uk/books?id=fU62CgAAQBAJ} {\emph{Modification}}.
\newblock Key Topics in Semantics and Pragmatics. Cambridge University Press.

\bibitem[{Muennighoff et~al.(2023)Muennighoff, Tazi, Magne, and Reimers}]{muennighoff2023mteb}
Niklas Muennighoff, Nouamane Tazi, Loic Magne, and Nils Reimers. 2023.
\newblock Mteb: Massive text embedding benchmark.
\newblock In \emph{Proceedings of the 17th Conference of the European Chapter of the Association for Computational Linguistics}, pages 2014--2037.

\bibitem[{Ni et~al.(2022)Ni, Abrego, Constant, Ma, Hall, Cer, and Yang}]{ni2022sentence}
Jianmo Ni, Gustavo~Hernandez Abrego, Noah Constant, Ji~Ma, Keith Hall, Daniel Cer, and Yinfei Yang. 2022.
\newblock Sentence-t5: Scalable sentence encoders from pre-trained text-to-text models.
\newblock In \emph{Findings of the Association for Computational Linguistics: ACL 2022}, pages 1864--1874.

\bibitem[{OpenAI(2024)}]{openai_embeddings}
OpenAI. 2024.
\newblock Embeddings.
\newblock \url{https://platform.openai.com/docs/guides/embeddings/}.

\bibitem[{Paperno and Baroni(2016)}]{paperno2016whole}
Denis Paperno and Marco Baroni. 2016.
\newblock When the whole is less than the sum of its parts: How composition affects pmi values in distributional semantic vectors.
\newblock \emph{Computational Linguistics}, 42(2):345--350.

\bibitem[{Paperno et~al.(2014)Paperno, Baroni et~al.}]{paperno2014practical}
Denis Paperno, Marco Baroni, et~al. 2014.
\newblock A practical and linguistically-motivated approach to compositional distributional semantics.
\newblock In \emph{Proceedings of the 52nd Annual Meeting of the Association for Computational Linguistics (Volume 1: Long Papers)}, pages 90--99.

\bibitem[{Pavlick and Callison-Burch(2016)}]{pavlick-callison-burch-2016-called}
Ellie Pavlick and Chris Callison-Burch. 2016.
\newblock \href {https://doi.org/10.18653/v1/S16-2014} {So-called non-subsective adjectives}.
\newblock In \emph{Proceedings of the Fifth Joint Conference on Lexical and Computational Semantics}, pages 114--119, Berlin, Germany. Association for Computational Linguistics.

\bibitem[{Pennington et~al.(2014)Pennington, Socher, and Manning}]{pennington2014glove}
Jeffrey Pennington, Richard Socher, and Christopher~D Manning. 2014.
\newblock Glove: Global vectors for word representation.
\newblock In \emph{Proceedings of the 2014 conference on empirical methods in natural language processing (EMNLP)}, pages 1532--1543.

\bibitem[{Pilehvar and Navigli(2015)}]{pilehvar2015senses}
Mohammad~Taher Pilehvar and Roberto Navigli. 2015.
\newblock From senses to texts: An all-in-one graph-based approach for measuring semantic similarity.
\newblock \emph{Artificial Intelligence}, 228:95--128.

\bibitem[{Purver et~al.(2021)Purver, Sadrzadeh, Kempson, Wijnholds, and Hough}]{purver2021incremental}
Matthew Purver, Mehrnoosh Sadrzadeh, Ruth Kempson, Gijs Wijnholds, and Julian Hough. 2021.
\newblock Incremental composition in distributional semantics.
\newblock \emph{Journal of Logic, Language and Information}, 30(2):379--406.

\bibitem[{Reimers and Gurevych(2019)}]{reimers2019sentence}
Nils Reimers and Iryna Gurevych. 2019.
\newblock Sentence-bert: Sentence embeddings using siamese bert-networks.
\newblock In \emph{Proceedings of the 2019 Conference on Empirical Methods in Natural Language Processing and the 9th International Joint Conference on Natural Language Processing (EMNLP-IJCNLP)}, pages 3982--3992.

\bibitem[{Ribeiro et~al.(2020)Ribeiro, Wu, Guestrin, and Singh}]{Ribeiro2020}
Marco~Tulio Ribeiro, Tongshuang Wu, Carlos Guestrin, and Sameer Singh. 2020.
\newblock \href {https://doi.org/10.18653/v1/2020.acl-main.442} {Beyond accuracy: Behavioral testing of {NLP} models with {C}heck{L}ist}.
\newblock In \emph{Proceedings of the 58th Annual Meeting of the Association for Computational Linguistics}, pages 4902--4912, Online. Association for Computational Linguistics.

\bibitem[{Sanh et~al.(2019)Sanh, Debut, Chaumond, and Wolf}]{Sanh2019DistilBERTAD}
Victor Sanh, Lysandre Debut, Julien Chaumond, and Thomas Wolf. 2019.
\newblock Distilbert, a distilled version of bert: smaller, faster, cheaper and lighter.

\bibitem[{Sommerauer and Fokkens(2019)}]{sommerauer2019conceptual}
Pia Sommerauer and Antske Fokkens. 2019.
\newblock Conceptual change and distributional semantic models: An exploratory study on pitfalls and possibilities.
\newblock In \emph{Proceedings of the 1st International Workshop on Computational Approaches to Historical Language Change}, pages 223--233.

\bibitem[{Sun and Zhou(2018)}]{Sun2018}
Liqun Sun and Zhi~Quan Zhou. 2018.
\newblock \href {https://doi.org/10.1109/ASWEC.2018.00021} {Metamorphic testing for machine translations: Mt4mt}.
\newblock In \emph{2018 25th Australasian Software Engineering Conference (ASWEC)}, pages 96--100.

\bibitem[{Tu et~al.(2021)Tu, Jiang, and Ding}]{Tu2021}
Kaiyi Tu, Mingyue Jiang, and Zuohua Ding. 2021.
\newblock \href {https://doi.org/10.3390/math9070726} {A metamorphic testing approach for assessing question answering systems}.
\newblock \emph{Mathematics}, 9(7).

\bibitem[{Vecchi et~al.(2013)Vecchi, Zamparelli, and Baroni}]{vecchi2013studying}
Eva~Maria Vecchi, Roberto Zamparelli, and Marco Baroni. 2013.
\newblock Studying the recursive behaviour of adjectival modification with compositional distributional semantics.
\newblock In \emph{Proceedings of the 2013 conference on empirical methods in natural language processing}, pages 141--151.

\bibitem[{Weir et~al.(2016)Weir, Weeds, Reffin, and Kober}]{weir2016aligning}
David Weir, Julie Weeds, Jeremy Reffin, and Thomas Kober. 2016.
\newblock Aligning packed dependency trees: a theory of composition for distributional semantics.
\newblock \emph{Computational Linguistics}, 42(4):727--761.

\end{thebibliography}

\appendix

\newpage


\section{List of adjectives and nouns} \label{sec:apd_adj_list}

\textbf{Subsective (Intersective):} wild, red, Canadian, depressed, square, seasonal, flamboyant, vigorous, loud, orange, shy.\\
\textit{Synonyms}: feral, crimson, North American, melancholic, cuboid, periodic, exuberant, robust, cacophonous, peach, timid.\\

\noindent
\textbf{Subsective (Non-Intersective):} skilful, powerful, particular, extreme, rare, unexpected.\\
\textit{Synonyms}: skilled, potent, specific, severe, uncommon, surprising.\\

\noindent
\textbf{Plain Non-Subsective:} former, alleged, apparent, arguable, assumed, believed, disputed, doubtful, erroneous, expected, faulty, future, historic, impossible, improbable, likely, ostensible, plausible, potential, proposed, putative, questionable, so-called, suspicious, theoretical, uncertain, unsuccessful.\\
\textit{Synonyms}: previous, suspected, seeming, debatable, presumed, assumed, doubted, dubious, mistaken, predicted, broken, upcoming, legendary, unachievable, unlikely, probable, apparent, possible, possible, suggested, supposed, dubious, commonly-named, dubious, philosophical, tentative, failed

\noindent
\textbf{Privative Non-Subsective:} artificial, counterfeit, deputy, ex-, fabricated, fictional, hypothetical, imaginary, mock, mythical, past, phony, spurious, virtual.\\
\textit{Synonyms}: fake, forged, vice, former, forged, fictitious, supposed, imagined, simulated, fantastical, prior, fake, bogus, simulated.\\

\noindent
\textbf{Ambiguous:} old, small, big.\\
\textit{Synonyms}: aged, tiny, large.\\

\noindent
\textbf{Nouns:} student, dog, potato, story, king, person, chair, occurence, law, problem, disaster, statement\\
\textit{Synonyms}: learner, canine, tater, narrative, monarch, human, seat, happening, regulation, difficulty, catastrophe, declaration.\\

\section{Full experimental results} \label{sec:apd_exp_results}

Tables~\ref{tab:full_set_dist_aan} and \ref{tab:full_setdist_pair_an} present the complete results of the set distance experiments for single phrases and phrase pairs, respectively (Section~\ref{sec:exp_intersect}). Table~\ref{tab:full_non-sub} presents the complete results for the phrase-word (non-subsectivity) distance experiment (Section~\ref{sec:exp_nonsubsect}).

\begin{table*}[ht!]
    \centering
    \begin{tabular}{@{}llllllllll@{}}
    \toprule
    \multicolumn{1}{c}{\textbf{Models}} & \multicolumn{9}{c}{\textbf{Adjective Type Pair}}                                                                                                                                                                                                                                                                                                                                                                                                                                                                                               \\ \midrule
    \multicolumn{1}{c}{\textbf{}}       & \begin{tabular}[c]{@{}l@{}}(S-I,\\ S-I)\end{tabular}      & \begin{tabular}[c]{@{}l@{}}(S-NI,\\ S-I)\end{tabular}   & \begin{tabular}[c]{@{}l@{}}(NS-Pl, \\ S-I)\end{tabular}  & \begin{tabular}[c]{@{}l@{}}(NS-Pr, \\ S-I)\end{tabular}  & \begin{tabular}[c]{@{}l@{}}(A,\\ S-I)\end{tabular}        & \begin{tabular}[c]{@{}l@{}}(S-I, \\ S-NI)\end{tabular} & \begin{tabular}[c]{@{}l@{}}(S-NI, \\ S-NI)\end{tabular} & \begin{tabular}[c]{@{}l@{}}(NS-Pl,\\ S-NI)\end{tabular} & \begin{tabular}[c]{@{}l@{}}(NS-Pr, \\ S-NI)\end{tabular}  \\ \midrule
    DPR                                 & 0.5242                                                    & 0.4268                                                  & 0.5314                                                   & 0.5222                                                   & 0.6288                                                    & 0.3939                                                 & 0.3694                                                  & 0.4347                                                  & 0.3839                                                    \\
    LaBSE                               & 0.9205                                                    & 0.9343                                                  & 0.9534                                                   & 0.9139                                                   & 0.9671                                                    & 0.9431                                                 & 0.9555                                                  & 0.9295                                                  & 0.9394                                                    \\
    Specter                             & 0.6667                                                    & 0.7273                                                  & 0.7177                                                   & 0.6661                                                   & 0.7348                                                    & 0.7955                                                 & 0.8861                                                  & 0.8184                                                  & 0.7857                                                    \\
    TE3-small                           & 0.9773                                                    & 0.9886                                                  & 0.9941                                                   & 0.9973                                                   & 0.9899                                                    & 0.9760                                                 & 0.9583                                                  & 0.9820                                                  & 0.9901                                                    \\
    NV-Embed-v2                         & 0.7788                                                    & 0.7134                                                  & 0.6838                                                   & 0.8139                                                   & 0.7475                                                    & 0.7437                                                 & 0.5583                                                  & 0.4892                                                  & 0.7123                                                    \\
    stella\_en\_1.5B\_v5                & 0.9992                                                    & 0.9949                                                  & 0.9992                                                   & 1.0                                                      & 1.0                                                       & 0.9962                                                 & 1.0                                                     & 0.9990                                                  & 0.9980                                                    \\ \midrule
    Glove                               & 1.0                                                       & 1.0                                                     & 1.0                                                      & 0.9393                                                   & 1.0                                                       & 1.0                                                    & 1.0                                                     & 1.0                                                     & 0.9414                                                    \\
    Word2Vec                            & 0.9969                                                    & 1.0                                                     & 0.9691                                                   & 0.9404                                                   & 1.0                                                       & 1.0                                                    & 1.0                                                     & 0.9686                                                  & 0.9394                                                    \\ \midrule
                                        & \begin{tabular}[c]{@{}l@{}}(A, \\ S-NI)\end{tabular}      & \begin{tabular}[c]{@{}l@{}}(S-I, \\ NS-Pl)\end{tabular} & \begin{tabular}[c]{@{}l@{}}(S-NI, \\ NS-Pl)\end{tabular} & \begin{tabular}[c]{@{}l@{}}(NS-Pl,\\ NS-Pl)\end{tabular} & \begin{tabular}[c]{@{}l@{}}(NS-Pr, \\ NS-Pl)\end{tabular} & \begin{tabular}[c]{@{}l@{}}(A,\\ NS-Pl)\end{tabular}   & \begin{tabular}[c]{@{}l@{}}(S-I, \\ NS-Pr)\end{tabular} & \begin{tabular}[c]{@{}l@{}}(S-NI,\\ NS-Pr)\end{tabular} & \begin{tabular}[c]{@{}l@{}}(NS-Pl, \\ NS-Pr)\end{tabular} \\ \midrule
    DPR                                 & 0.5324                                                    & 0.4571                                                  & 0.3529                                                   & 0.3681                                                   & 0.3997                                                    & 0.5401                                                 & 0.5536                                                  & 0.4345                                                  & 0.4782                                                    \\
    LaBSE                               & 0.9814                                                    & 0.9584                                                  & 0.9274                                                   & 0.8360                                                   & 0.8858                                                    & 0.9588                                                 & 0.8874                                                  & 0.9176                                                  & 0.8536                                                    \\                                    
    Specter                             & 0.8194                                                    & 0.7932                                                  & 0.8009                                                   & 0.6899                                                   & 0.7235                                                    & 0.7963                                                 & 0.7002                                                  & 0.7380                                                  & 0.6960                                                    \\
    TE3-small                           & 0.9769                                                    & 0.9823                                                  & 0.9784                                                   & 0.9160                                                   & 0.9625                                                    & 0.9609                                                 & 0.9729                                                  & 0.9692                                                  & 0.9372                                                    \\
    NV-Embed-v2                         & 0.7222                                                    & 0.6512                                                  & 0.4789                                                   & 0.3251                                                   & 0.5223                                                    & 0.5895                                                 & 0.8236                                                  & 0.7292                                                  & 0.5686                                                    \\
    stella\_en\_1.5B\_v5                & 1.0                                                       & 0.9992                                                  & 0.9974                                                   & 0.9771                                                   & 0.9894                                                    & 1.0                                                    & 0.9984                                                  & 0.9931                                                  & 0.9850                                                    \\ \midrule
    Glove                               & 1.0                                                       & 1.0                                                     & 1.0                                                      & 0.9992                                                   & 0.9442                                                    & 1.0                                                    & 0.9393                                                  & 0.9414                                                  & 0.9442                                                    \\
    Word2Vec                            & 1.0                                                       & 0.9691                                                  & 0.9404                                                   & 0.9394                                                   & 0.9122                                                    & 0.9691                                                 & 0.9404                                                  & 0.9394                                                  & 0.9122                                                    \\ \midrule
                                        & \begin{tabular}[c]{@{}l@{}}(NS-Pr, \\ NS-Pr)\end{tabular} & \begin{tabular}[c]{@{}l@{}}(A,\\ NS-Pr)\end{tabular}    & \begin{tabular}[c]{@{}l@{}}(S-I,\\ A)\end{tabular}       & \begin{tabular}[c]{@{}l@{}}(S-NI,\\ A)\end{tabular}      & \begin{tabular}[c]{@{}l@{}}(NS-Pl,\\ A)\end{tabular}      & \begin{tabular}[c]{@{}l@{}}NS-Pr, \\ A)\end{tabular}   & \begin{tabular}[c]{@{}l@{}}(A,\\ A)\end{tabular}        &                                                         &                                                           \\ \midrule
    DPR                                 & 0.5069                                                    & 0.6409                                                  & 0.6263                                                   & 0.5278                                                   & 0.6029                                                    & 0.5734                                                 & 0.5139                                                  &                                                         &                                                           \\
    LaBSE                               & 0.6483                                                    & 0.9166                                                  & 0.9848                                                   & 0.9861                                                   & 0.9701                                                    & 0.9523                                                 & 1.0                                                     &                                                         &                                                           \\
    Specter                             & 0.6662                                                    & 0.7619                                                  & 0.7753                                                   & 0.8519                                                   & 0.7891                                                    & 0.7361                                                 & 0.6806                                                  &                                                         &                                                           \\
    TE3-small                           & 0.9139                                                    & 0.9583                                                  & 0.9798                                                   & 0.9491                                                   & 0.9856                                                    & 0.9722                                                 & 0.9722                                                  &                                                         &                                                           \\
    NV-Embed-v2                         & 0.6346                                                    & 0.7679                                                  & 0.7879                                                   & 0.7639                                                   & 0.7006                                                    & 0.8115                                                 & 0.4861                                                  &                                                         &                                                           \\
    stella\_en\_1.5B\_v5                & 0.9547                                                    & 0.9921                                                  & 1.0                                                      & 0.9954                                                   & 1.0                                                       & 1.0                                                    & 1.0                                                     &                                                         &                                                           \\ \midrule
    Glove                               & 0.8873                                                    & 0.9345                                                  & 1.0                                                      & 1.0                                                      & 1.0                                                       & 0.9345                                                 & 1.0                                                     &                                                         &                                                           \\
    Word2Vec                            & 0.8736                                                    & 0.9404                                                  & 1.0                                                      & 1.0                                                      & 0.9691                                                    & 0.9404                                                 & 1.0                                                     &                                                         &                                                           \\ \bottomrule \\
    \end{tabular}
    \caption{Satisfaction score (consistency) for the set distance property (Equation~\ref{eq:expect_intersect}), for noun phrases with pairs of adjectives of the indicated types (AAN format). We use the following shorthand notation in the table columns:
    A: Ambiguous, S-I: Subsective-Intersective, S-NI: Subsective Non-Intersective, NS-Pl: Plain Non-Subsective, NS-Pr: Privative Non-Subsective}
    \label{tab:full_set_dist_aan}
    \end{table*}
    
\begin{table*}[h!]
    \centering
    \begin{tabular}{@{}llllllllll@{}}
    \toprule
    \multicolumn{1}{c}{\textbf{Models}} & \multicolumn{9}{c}{\textbf{Adjective Type Pair}}                                                                                                                                                                                                                                                                                                                                                                                                                                                                                               \\ \midrule
    \multicolumn{1}{c}{\textbf{}}       & \begin{tabular}[c]{@{}l@{}}(S-I,\\ S-I)\end{tabular}      & \begin{tabular}[c]{@{}l@{}}(S-I,\\ S-NI)\end{tabular}   & \begin{tabular}[c]{@{}l@{}}(S-I, \\ NS-Pl)\end{tabular}  & \begin{tabular}[c]{@{}l@{}}(S-I, \\ NS-Pr)\end{tabular}  & \begin{tabular}[c]{@{}l@{}}(S-I,\\ A)\end{tabular}        & \begin{tabular}[c]{@{}l@{}}(S-NI, \\ S-NI)\end{tabular} & \begin{tabular}[c]{@{}l@{}}(S-NI, \\ NS-Pl)\end{tabular} & \begin{tabular}[c]{@{}l@{}}(S-NI,\\ NS-Pr)\end{tabular} & \begin{tabular}[c]{@{}l@{}}(S-NI, \\ A)\end{tabular}  \\ \midrule
    DPR                                 & 0.5000                                                    & 0.3216                                                  & 0.3380                                                   & 0.4999                                                   & 0.4238                                                    & 0.5000                                                  & 0.4882                                                   & 0.6695                                                  & 0.5834                                                    \\
    LaBSE                               & 0.5000                                                    & 0.4252                                                  & 0.3386                                                   & 0.5268                                                   & 0.3316                                                    & 0.5000                                                  & 0.4155                                                   & 0.6216                                                  & 0.3877                                                    \\
    Specter                             & 0.5000                                                    & 0.6530                                                  & 0.5461                                                   & 0.5042                                                   & 0.5735                                                    & 0.5000                                                  & 0.4091                                                   & 0.3710                                                  & 0.4014                                                    \\
    TE3-small                           & 0.5000                                                    & 0.5108                                                  & 0.4824                                                   & 0.4822                                                   & 0.8223                                                    & 0.5000                                                  & 0.4527                                                   & 0.4552                                                  & 0.8723                                                    \\
    NV-Embed-v2                         & 0.5000                                                    & 0.5412                                                  & 0.5148                                                   & 0.5158                                                   & 0.8230                                                    & 0.5000                                                  & 0.4769                                                   & 0.4825                                                  & 0.8759                                                    \\
    stella\_en\_1.5B\_v5                & 0.5000                                                    & 0.7597                                                  & 0.6379                                                   & 0.5782                                                   & 0.9090                                                    & 0.5000                                                  & 0.3897                                                   & 0.3241                                                  & 0.6614                                                    \\ \midrule
    Glove                               & 0.5000                                                    & 0.6565                                                  & 0.6940                                                   & 0.7022                                                   & 0.4684                                                    & 0.5000                                                  & 0.5428                                                   & 0.5552                                                  & 0.3215                                                    \\
    Word2Vec                            & 0.5000                                                    & 0.7536                                                  & 0.6518                                                   & 0.4879                                                   & 0.9982                                                    & 0.5000                                                  & 0.4289                                                   & 0.2953                                                  & 0.8103                                                    \\ \midrule
                                        & \begin{tabular}[c]{@{}l@{}}(NS-Pl, \\ NS-Pl)\end{tabular}      & \begin{tabular}[c]{@{}l@{}}(NS-Pl, \\ NS-Pr)\end{tabular} & \begin{tabular}[c]{@{}l@{}}(NS-Pl, \\ A)\end{tabular} & \begin{tabular}[c]{@{}l@{}}(NS-Pr,\\ NS-Pr)\end{tabular} & \begin{tabular}[c]{@{}l@{}}(NS-Pr, \\ A)\end{tabular} & \begin{tabular}[c]{@{}l@{}}(A,\\ A)\end{tabular}   \\ \midrule
    DPR                                 & 0.5000                                                         & 0.6656                                                    & 0.5971                                                & 0.5000                                                   & 0.4487                                                & 0.5000                                                 \\
    LaBSE                               & 0.5000                                                         & 0.6838                                                    & 0.5209                                                & 0.5000                                                   & 0.2902                                                & 0.5000                                                 \\
    Specter                             & 0.5000                                                         & 0.4595                                                    & 0.5145                                                & 0.5000                                                   & 0.5318                                                & 0.5000                                                 \\
    TE3-small                           & 0.5000                                                         & 0.5001                                                    & 0.8100                                                & 0.5000                                                   & 0.8334                                                & 0.5000                                                 \\
    NV-Embed-v2                         & 0.5000                                                         & 0.5016                                                    & 0.8686                                                & 0.5000                                                   & 0.9053                                                & 0.5000                                                 \\
    stella\_en\_1.5B\_v5                & 0.5000                                                         & 0.4414                                                    & 0.7617                                                & 0.5000                                                   & 0.8252                                                & 0.5000                                                 \\ \midrule
    Glove                               & 0.5000                                                         & 0.5087                                                    & 0.2626                                                & 0.5000                                                   & 0.2558                                                & 0.5000                                                 \\
    Word2Vec                            & 0.5000                                                         & 0.3817                                                    & 0.8503                                                & 0.5000                                                   & 0.8642                                                & 0.5000                                                 \\ \bottomrule \\
    \end{tabular}
    \caption{Satisfaction score (consistency) for the set distance property across adjective phrase pairs (Equation~\ref{eq:pair_set_dist}), for noun phrases with pairs of adjectives of the indicated types (AN format). We use the following shorthand notation in the table columns:
    A: Ambiguous, S-I: Subsective-Intersective, S-NI: Subsective Non-Intersective, NS-Pl: Plain Non-Subsective, NS-Pr: Privative Non-Subsective}
    \label{tab:full_setdist_pair_an}
    \end{table*}

\begin{table*}[ht!]
\resizebox{\textwidth}{!}{
\begin{tabular}{@{}lp{2cm}p{2.3cm}p{2.4cm}p{2.4cm}l@{}}
\toprule
\multicolumn{1}{c}{\multirow{2}{*}{\textbf{Models}}} & \multicolumn{5}{c}{\textbf{Adjective Type}}                                                                                                                                                                                                                                                                                                                                                                 \\
\multicolumn{1}{c}{}                                 & \multicolumn{1}{c}{\begin{tabular}[c]{@{}c@{}}Subsective \\ (Intersective)\end{tabular}} & \multicolumn{1}{c}{\begin{tabular}[c]{@{}c@{}}Subsective \\ (Non-Intersective)\end{tabular}} & \multicolumn{1}{c}{\begin{tabular}[c]{@{}c@{}}Non-Subsective \\ (Plain)\end{tabular}} & \multicolumn{1}{c}{\begin{tabular}[c]{@{}c@{}}Non-Subsective \\ (Privative)\end{tabular}} & \multicolumn{1}{c}{Ambiguous} \\ \midrule
DPR                                                  & 0.4621                                                                                   & 0.3750                                                                                       & 0.4784                                                                                & 0.5357                                                                                    & 0.3889       \\
LaBSE                                                & 0.3560                                                                                   & 0.3055                                                                                       & 0.5123                                                                                & 0.3273                                                                                    & 0.1944       \\
Specter                                              & 0.4848                                                                                   & 0.3056                                                                                       & 0.4907                                                                                & 0.5714                                                                                    & 0.3333   \\
TE3-small                                            & 0.8106                                                                                   & 0.7500                                                                                       & 0.7438                                                                                & 0.7738                                                                                    & 0.3889  \\
NV-Embed-v2                                          & 0.8409                                                                                   & 0.7917                                                                                       & 0.7901                                                                                & 0.8274                                                                                    & 0.8056   \\ 
stella\_en\_1.5B\_v5                                 & 0.8106                                                                                   & 0.5556                                                                                       & 0.5772                                                                                & 0.6369                                                                                    & 0.3333       \\ \midrule
Glove                                                & 0.6060                                                                                   & 0.2222                                                                                       & 0.2191                                                                                &  0.3214                                                                                   & 0.2777      \\
Word2Vec                                             & 0.5530                                                                                   & 0.2083                                                                                       &  0.3364                                                                               & 0.4940                                                                                    & 0.0       \\ \bottomrule \\
\end{tabular}%
}
\caption{Non-subsectivity experiment, reporting satisfaction score (consistency) for the property in equation \ref{eq:emb_rel_ni}, as its expectation for the phrase dataset (Equation~\ref{eq:expect_nonsubsect}).}
\label{tab:full_non-sub}
\end{table*}

\end{document}